\documentclass[runningheads]{llncs}

\usepackage{eccv}

\usepackage{eccvabbrv}

\usepackage{graphicx}
\usepackage{booktabs}

\usepackage[accsupp]{axessibility}  

\usepackage{graphicx}
\usepackage{multirow}
\usepackage{wrapfig}
\usepackage{amsmath}
\usepackage{caption}
\usepackage{subcaption}
\usepackage{enumitem}
\usepackage{graphicx}
\usepackage{caption}
\usepackage{color}
\usepackage[most]{tcolorbox}
\usepackage{lipsum}
\usepackage[symbol]{footmisc}
\usepackage[dvipsnames]{xcolor}
\usepackage{colortbl}
\usepackage{color}

\usepackage{graphicx}
\usepackage{wrapfig}

\usepackage{arydshln}

\usepackage{adjustbox}
\usepackage[misc]{ifsym}

\usepackage[dvipsnames]{xcolor}

\usepackage[pagebackref,breaklinks,colorlinks,citecolor=eccvblue]{hyperref}
\usepackage{hyperref}

\usepackage{orcidlink}

\begin{document}

\title{Efficient 3D-Aware Facial Image Editing via
Attribute-Specific Prompt Learning} 

\titlerunning{Efficient 3D-Aware Facial Image Editing}


\author{Amandeep Kumar\thanks{Authors contributed equally. \\ More details are available at \href{https://awaisrauf.github.io/3d_face_editing}{\url{awaisrauf.github.io/3d_face_editing}}.
}
\textsuperscript{1}  \quad Muhammad Awais$^\ast$\textsuperscript{1} \quad Sanath Narayan\textsuperscript{2} \\ \quad Hisham Cholakkal\textsuperscript{1}  \quad Salman Khan\textsuperscript{1} \quad Rao Muhammad Anwer\textsuperscript{1}  \vspace{.1cm}}
\authorrunning{Kumar et al.}

\institute{Mohamed bin Zayed University of Artificial Intelligence, UAE  \and
Technology Innovation Institute, UAE}

\maketitle

\begin{center}
        \centering
        \captionsetup{type=figure}
        {\includegraphics[width=0.9\textwidth]{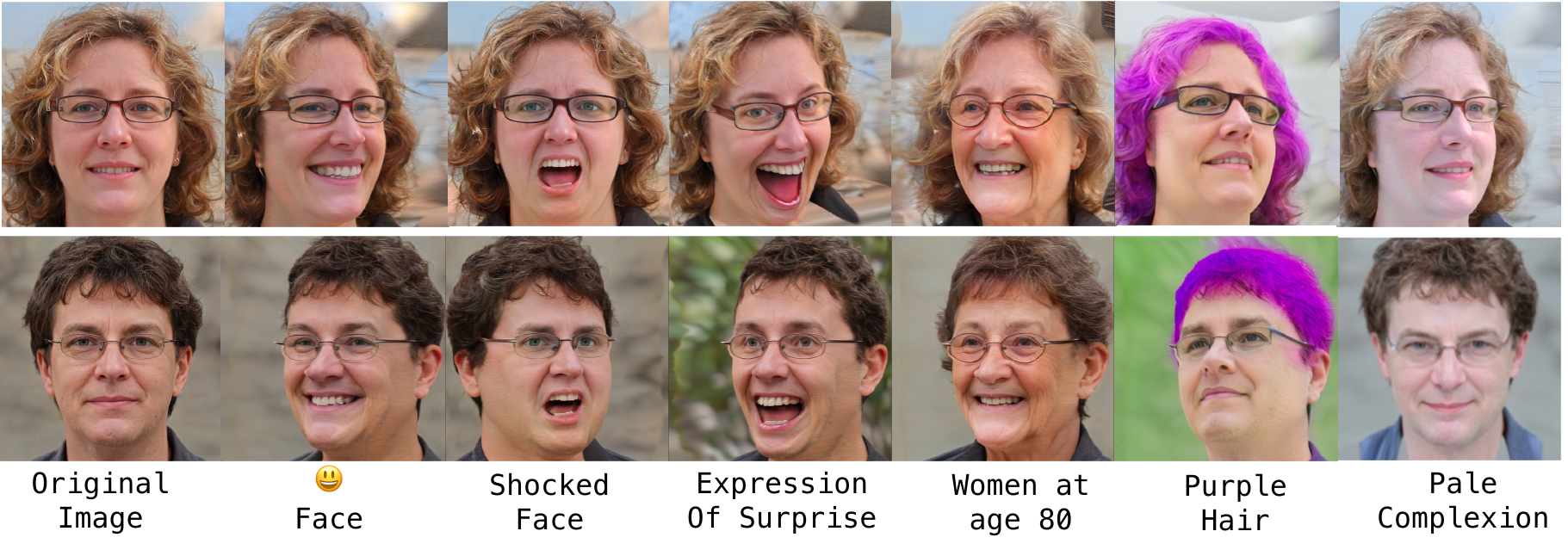}}\vspace{-0.3cm}
        \caption{Examples of text-driven and 3D consistent attribute editing using our method. The first column displays the original image, while subsequent columns depict attribute editing guided by text prompts at randomly sampled target angles.}
        \label{fig:intro-image}
        \vspace{-0.7em}
    \end{center}


\begin{abstract}
Drawing upon StyleGAN's expressivity and disentangled latent space, existing 2D approaches employ textual prompting to edit facial images with different attributes. In contrast, 3D-aware approaches that generate faces at different target poses require attribute-specific classifiers, learning separate model weights for each attribute, and are not scalable for novel attributes. In this work, we propose an efficient, plug-and-play, 3D-aware face editing framework based on attribute-specific prompt learning, enabling the generation of facial images with controllable attributes across various target poses. To this end, we introduce a text-driven learnable style token-based latent attribute editor (LAE). The LAE harnesses a pre-trained vision-language model to find text-guided attribute-specific editing direction in the latent space of any pre-trained 3D-aware GAN. It utilizes learnable style tokens and style mappers to learn and transform this editing direction to 3D latent space. To train LAE with multiple attributes, we use directional contrastive loss and style token loss.
Furthermore, to ensure view consistency and identity preservation across different poses and attributes, we employ several 3D-aware identity and pose preservation losses. Our experiments show that our proposed framework generates high-quality images with 3D awareness and view consistency while maintaining attribute-specific features. We demonstrate the effectiveness of our method on different facial attributes, including hair color and style, expression, and others. 
\end{abstract}

\section{Introduction}
\label{sec:intro}

StyleGAN~\cite{karras2019style, karras2020analyzing} has demonstrated exceptional capabilities in generating unconditional, photorealistic 2D images. The disentangled properties of the learned latent space of StyleGAN have facilitated attribute-specific realistic image editing by finding and modifying semantic editing directions ~\cite{abdal2021styleflow, shen2020interpreting,harkonen2020ganspace,tewari2020stylerig}.
\begin{wraptable}{r}{0.6\textwidth}
    \vspace{-30pt}
    \caption{Comparison of our method's capabilities with existing methods. Attrs. here denote attributes. }
    \resizebox{0.6\textwidth}{!}{%
    \begin{tabular}{l|cccc}
    \toprule
          \multirow{2}{*}{Method}   &   \multirow{2}{*}{Editability}     & 3D & Text &  Efficient for   \\
                   &         & Aware & Driven   &  Novel Attrs.  \\ 
        \toprule
        StyleGAN~\cite{karras2019style}  & \textcolor{OliveGreen}{\textbf{$\checkmark$}}  & \textcolor{red}{\textbf{ $\times$}}  & \textcolor{red}{\textbf{ $\times$}} & \textcolor{red}{\textbf{ $\times$}} \\
        CLIPStyle~\cite{patashnik2021styleclip} &  \textcolor{OliveGreen}{\textbf{$\checkmark$}} &  \textcolor{red}{\textbf{ $\times$}} & \textcolor{OliveGreen}{\textbf{$\checkmark$}} &  \textcolor{red}{\textbf{ $\times$}} \\
        EG3D~\cite{chan2022efficient}      &   \textcolor{red}{\textbf{ $\times$}} & \textcolor{OliveGreen}{\textbf{$\checkmark$}}  &  \textcolor{red}{\textbf{ $\times$}} &  \textcolor{red}{\textbf{ $\times$}} \\
        GMPI~\cite{zhao2022generative}      &   \textcolor{red}{\textbf{ $\times$}} & \textcolor{OliveGreen}{\textbf{$\checkmark$}}  &  \textcolor{red}{\textbf{ $\times$}} &  \textcolor{red}{\textbf{ $\times$}} \\
        PREIM3D~\cite{zhao2022generative}   & \textcolor{OliveGreen}{\textbf{$\checkmark$}} & \textcolor{OliveGreen}{\textbf{$\checkmark$}} & \textcolor{red}{\textbf{ $\times$}} & \textcolor{red}{\textbf{ $\times$}}  \\
        \cellcolor{cyan!25}Ours      & \cellcolor{cyan!25} \textcolor{OliveGreen}{\textbf{$\checkmark$}} & \cellcolor{cyan!25} \textcolor{OliveGreen}{\textbf{$\checkmark$}} & \cellcolor{cyan!25} \textcolor{OliveGreen}{\textbf{$\checkmark$}} &
        \cellcolor{cyan!25} \textcolor{OliveGreen}{\textbf{$\checkmark$}} \\
         \bottomrule
    \end{tabular}}

    \label{tab:summary}
    \vspace{-15pt}

\end{wraptable}
Furthermore, recent studies have harnessed foundational vision-language models (e.g., CLIP ~\cite{radford2021learning}) to facilitate text-guided attribute editing~\cite{patashnik2021styleclip}.

Recently, StyleGAN-based architecture has also been adapted for 3D-aware and view-consistent image generation~\cite{zhao2022generative, chan2022efficient, gu2021stylenerf, lin20223d, sun2022ide}. For instance, GMPI~\cite{zhao2022generative} has modified 2D StyleGAN by introducing an alpha branch to learn alpha maps to generate 3D-aware multiplane images efficiently. StyleNeRF~\cite{gu2021stylenerf} integrates the neural radiance field (NeRF) into a style-based generator and EG3D \cite{chan2022efficient} which incorporate triplane representation along with style mapping network to generate view consistent images.   However, compared to 2D images, achieving 3D-aware image editing encounters several challenges, as it requires not only maintaining consistency between the edited and original images but also ensuring view consistency across different poses while preserving the facial identity.

Building on 3D GANs, some works have introduced attribute editing methods that can manipulate a set of attributes in 3D~\cite{lin20223d, sun2022ide, roich2022pivotal, li2023preim3d}. However, these methods are limited to editing a predetermined set of attributes as they require training attribute classifiers for every new attribute. This training takes several hours on large datasets of training for every attribute. Moreover, these methods struggle with maintaining identity and view consistency across a wide variety of camera angles. Finally, due to the use of pre-trained classifiers, novel attribute editing is expensive and limited. A more natural way is to use textual prompts that not only significantly extend the range of manipulations but also provide a more intuitive form of interaction that closely resembles humans \cite{patashnik2021styleclip}.

In this work, we introduce a plug-and-play module designed to efficiently integrate text-guided editing of novel attributes into 3D-aware GANs while preserving both the 3D pose and identity.
Our proposed module, Learned Attribute Editor (LAE), finds semantic editing directions specified by a natural language prompt, thereby facilitating text-guided manipulation of novel attributes. The language prompt specifying the target attribute is concatenated with learnable style tokens and passed through a frozen CLIP text encoder. These prompt encodings are then passed through a set of style mappers that learn to transform text-specified modifications to the latent space of 3D-aware GAN. The combination of learnable style tokens and vision-language models makes the editing process extremely efficient (from hours to a few minutes), as LAE does not need to train classifiers on large datasets for every new attribute. 

Training the LAE module with language supervision poses challenges, as standard CLIP loss~\cite{radford2021learning} may struggle to converge in the 3D latent space. 
To overcome this issue, we introduce a 3D-aware DCLIP loss~\cite{Gal2021StyleGANNADACD} which utilizes multi-view images generated on-the-fly. To improve the view consistency and preserve multi-view identity, we introduce novel losses, including style token contrastive loss, IDVC, and several 3D-aware pose and identity preservation losses. These losses enable maintaining the distinctive characteristics of each attribute, aligning the CLIP-space direction with learnable text features and the multi-view generated images, and ensuring consistency across different camera poses. We primarily employ GMPI~\cite{zhao2022generative}, which adapts StyleGAN to generate color images and corresponding alpha maps. We also demonstrate our method's plug-and-play nature by integrating it with several different 3D-aware GANs.
Our experimental results validate the efficacy of our proposed method, showcasing its superior generative performance in both qualitative and quantitative ways. We present a representative sample of editing made by our method at random camera poses in Figure~\ref{fig:intro-image}.

\section{Related Works}

\textbf{Text Guided Image Manipulation in 2D. }
Generative adversarial networks (GANs)\cite{goodfellow2020generative} have demonstrated remarkable performance in generating realistic, unconditional 2D images\cite{karras2017progressive, radford2015unsupervised, brock2018large, miyato2018spectral}. Among these models, style-based GAN (StyleGAN) has demonstrated state-of-the-art performance~\cite{karras2019style, karras2020analyzing, karras2020training}. Beyond their ability in realistic photo generation, StyleGAN's latent space exhibits disentanglement properties~\cite{collins2020editing, shen2020interpreting, harkonen2020ganspace, tewari2020stylerig, wu2021stylespace}. These properties allow leveraging pre-trained models for a wide array of manipulations, such as altering hair color or changing emotions, achieved by traversing specific directions identified through manual examination~\cite{harkonen2020ganspace, shen2020interpreting, wu2021stylespace}, or attribute classifiers~\cite{shen2020interfacegan, abdal2021styleflow}. 


Recently, a considerable effort has been dedicated to cross-modal vision-language (VL) representation learning, and models have shown impressive performance across a wide variety of tasks~\cite{radford2021learning, yuan2021florence, jia2021scaling, yao2021filip,singh2022flava, li2022blip, alayrac2022flamingo}. Specifically, Contrastive Language-Image Pre-training (CLIP) ~\cite{radford2021learning}, trained on 400 million image-text pairs. CLIP's learned embeddings have shown to be extremely powerful across domains. StyleCLIP~\cite{patashnik2021styleclip} leverage pre-trained CLIP model to find manipulation direction via. text prompts. Several subsequent works have also explored text-guided image manipulation in 2D. However, our work is different as it delves into 3D space, which presents greater challenges due to additional constraints involving view consistency and 3D awareness. Furthermore, our method showcases significantly improved efficiency, which is attributed to our proposed style tokens.

\noindent\textbf{Generative 3D-aware image synthesis and manipulation. }
Extending the capabilities of 2D GANs to 3D settings has gained significant attention recently. These methods rely on a combination of 3D-structure aware inductive bias in the generator and utilization of neural rendering engines to get view-consistent results. These approaches include Mesh-based appraoches~\cite{liao2020towards, szabo2019unsupervised}, Voxel-based GANs~\cite{gadelha20173d, henzler2019escaping, nguyen2020blockgan, wu2016learning, zhu2018visual}, but may have limited expressiveness and high memory and computation requirements, respectively. Moreover, fully implicit representations-based approaches ~\cite{chan2021pi, schwarz2020graf} have also been proposed, but their slow querying and sampling make them difficult to use in training.  Several works have also proposed that utilize a hybrid approach ~\cite{xie2023high, chan2022efficient, gu2021stylenerf, xu20223d, kumar2023generative}. 

However, in this work, we utilize multiple 3D aware models like GMPI, which uses multiplane images (MPIs)~\cite{zhou2018stereo} for image representation and adapts a StyleGAN to generate unconditional 3D-aware generation, EG3D \cite{chan2022efficient} and others. Our method aims to edit image attributes via. text while keeping 3D view-consistency. The closest work to our proposed method is PREIMD3D~\cite{li2023preim3d}, which is based on EG3D~\cite{chan2022efficient} for 3D image generation and finds a semantic edit direction in the inversion manifold such that the corresponding EG3D generated image flips the binary label of an attribute-specific pre-trained classifier. However, these method requires more resources to train for novel attributes.

\begin{figure*}[t!]
\centering
\scalebox{0.9}{
\includegraphics[width=0.97\textwidth]{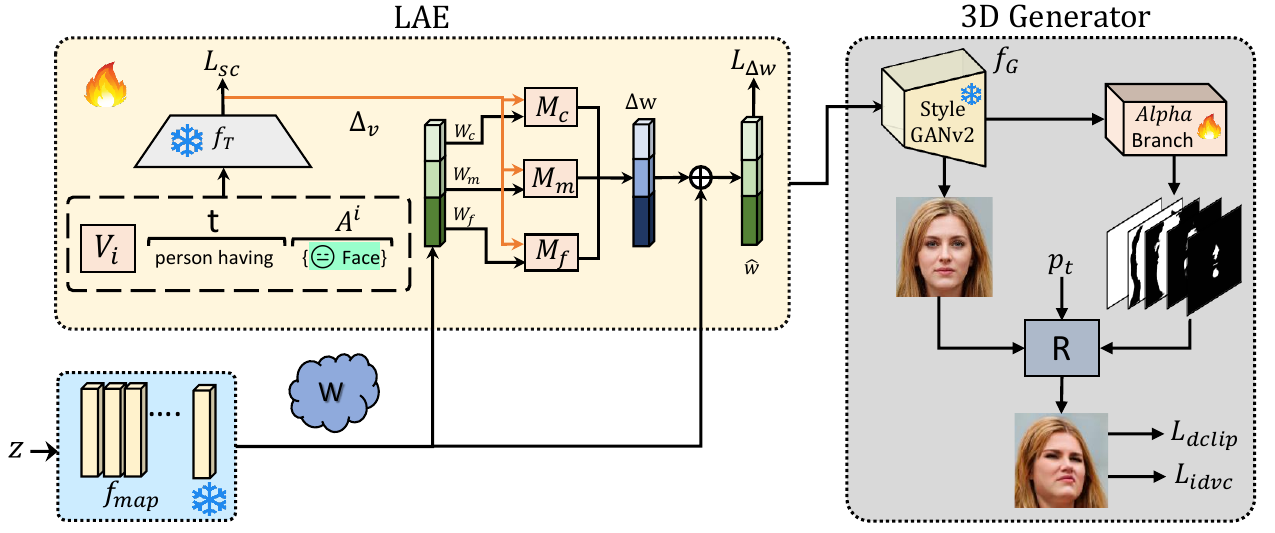}}
\caption{The overall architecture of our proposed method is based on attribute-specific prompt learning.  
Our proposed framework comprises a learnable prompt-based latent attribute editor (LAE), a mapping network $f_{map}$, RGB$\alpha$ generator $f_G$ along with a differentiable renderer $R$. The LAE consists of learnable style tokens, a CLIP-based text encoder $f_T$, and style mappers ($M$). The input-to-text encoder is a textual prompt for $i$-th attribute, which consists of textual prompt $A^i$, a textual instruction $t$, and a learnable token $V_i$. Text encoder converts this to $\Delta v$, which is then fed to style mappers ($M$). Style mappers map $\Delta v$ to the latent space of StyleGAN, which produces RGB images along with alpha maps. These alpha maps, along with target pose $p_t$, are then fed to a renderer that generates an image at a given pose. 
We introduce a learnable prompt-based attribute editor LAE that enables facial image generation with controllable attributes at different target poses within a single framework. }
\label{fig:method}
\vspace{-15pt}
\end{figure*}

\section{Method}
\label{sec:methodoogy}
In this work, our goal is to leverage a 3D GAN to enable novel attribute editing driven by natural language prompts that are 3D-aware and view-consistent. We use frozen 3D-aware StyleGAN called GMPI~\cite{zhao2022generative} to edit 3D images and guide its latent space ($\mathcal{W}$) via. our proposed latent attribute editor (LAE) in the target attribute direction by utilizing CLIP~\cite{radford2021learning}). Our method is significantly efficient as it only requires training of attribute-specific style tokens and linear-layer-based style mappers, which can be plugged in with different 3D generative models.

\subsection{Problem Formulation}
\label{sec:problem_formulation}
Given an input latent code $z$, an attribute editing instruction $A^i$ and a target camera pose $p_{t}$, our goal is to generate multiplane representations $\mathcal{M}$ that can be used to render 3D-aware and view-consistent images having the given attributes. A multiplane image can be represented as $(C_i, \alpha_i, d_i)$ for $L$ fronto-parallel planes, where $C \in \mathbb{R}^{H\times H \times 3}$ denotes color texture, $\alpha_i \in [0, 1]^{H\times H \times 1}$ and $d_i \in \mathbb{R}$ denotes alpha maps and depth for corresponding plane (distance from a camera). We use GMPI~\cite{zhao2022generative}, which simplifies this task to the generation of a single color texture image across all planes along with alpha maps. The alpha maps along with color texture $C_{i}$ are then fed to a renderer $R$ to generate an image at a specific camera angle. However, we also want to edit these images to have an arbitrary attribute $A^i$ specified by natural language text. Hence, the goal of our generator $f_G$ is to generate an RGB image $C$ with a particular attribute and corresponding alpha maps given a latent code $z$, a textual instruction $A^i$ and depth of the planes $d_i$: $M = \{C, \{ \alpha_i, ..., \alpha_L\}\} = f_G (z, A^i, \{ d_1, ..., d_L \})$. 

\subsection{Overview of Proposed Pipeline} 
\label{sec:overview}

Recent 3D face editing methods mostly suffer from limitations such as rigidity and time consumption, primarily because they rely on predefined attribute classes. This dependence requires training pre-trained attribute classifiers on large datasets, which can be computationally intensive. As a result, these methods struggle to adapt to novel attributes in real-time scenarios, limiting their practical utility for 3D-aware editing. To address these challenges, we introduce a learnable prompt-based attribute editor $LAE$ module within the GMPI framework and 3D-aware attribute editing, identity, and pose preservation losses for synthesizing and editing face images with prompt controllable attributes (\eg, hair color, style, expressions) while maintaining the view-consistency across various target poses. Further, the proposed $LAE$ module can also be integrated into other state-of-art 3D generation methods, enabling the editing capabilities while maintaining the identity and the view consistency across multiple camera poses.

 Figure~\ref{fig:method} presents our overall framework comprising a mapping network $f_{map}$, language-driven attribute editor $LAE$, RGB image ($C$) and $\alpha$-maps generator $f_G$ and a differentiable renderer $R$. Given a noise code $z \sim \mathcal{N}(0, 1)$, the mapping network maps it to the latent code $w \in \mathcal{W}$.
 Further, this latent code $w$ is edited within the $LAE$ module by input prompt $P_A^i$ (a combination of attribute prompt $A^i$, system prompt $t$, and learnable style tokens $V_i$). In particular, within the $LAE$, attribute-specific tokens $P_A^i$ are learned and mapped into textual embeddings $f_T$ using a text encoder $f_T(.)$. The resulting $f_T$ and latent code $w$ are then utilized to obtain the edited latent code $\hat{{w}}$ using style mappers $M_c$, $M_m$, and $M_f$. The edited $\hat{w}$ output by $LAE$ is subsequently fed into  RGB$\alpha$ generator $f_G$, which generates RGB image and alpha maps, which are then fed to the renderer $R$ for synthesizing images with the desired attributes at specified target poses $p_t$. 

Our proposed framework is trained end-to-end using the proposed 3D-aware attribute editing loss, prompt-based editing loss, and 3D-aware identity and pose preservation losses. While prompt-based editing loss ($\mathcal{L}_{dclip}$, $\mathcal{L}_{sc}$) enables controllable editing of attributes in the generated face images, the identity preservation loss ($\mathcal{L}_{id}$, $\mathcal{L}_{idvc}$, $\mathcal{L}_{latent}$, and $\mathcal{L}_{\alpha}$) strives to maintain the identity and camera pose. In the following, we first explain our language-driven attribute editor (LAE) and then losses that are designed to preserve the identity and 3D consistency of the generated images.

\subsection{Latent Attribute Editor (LAE) with Text Driven Editing}
\label{sec:LAE}

While the latent space of StyleGAN~\cite{karras2019style} has shown to be fairly disentangled, it still requires finding editing direction corresponding to an attribute. Since our goal is text-driven editing, an important question is: how can we effectively learn and extract information from the attribute prompt $A_i$ so that the network can generate images with the desired attributes? Inspired by \cite{patashnik2021styleclip}, we propose a text-driven latent attribute editor (LAE), which consists of style tokens and style mappers and is trained on Directional Clip and style token contrastive losses.

\noindent \textbf{Style Tokens and Style Mappers. }
In contrast to the existing works, we design a general prompt $P^i_A$ to represent the given attributes $[A]$. $P$ consists of learnable prompt vectors $\{[V]^{i}_{1}, [V]^{i}_{2}, \cdots, [V]^{i}_{m}\}$ and embedding of the attribute prompt  $[A^i]$. The system prompts $t$, which are independent of each class,
\begin{align*}
\mathbf{P}^i_A = [V]^{i}_{1}, [V]^{i}_{2}, \cdots, [V]^{i}_{m}, [t]_1, [t]_2, \cdots, [t]_l [A], 
\end{align*}
where $ \{\{[V]^{i}_{j} \in \mathbb{R}^{d_l} \}_{j=1}^m\}_{i=1}^n$, $n$ is the number of attributes and $m$ is number of learnable prompts, $\{t_l|_{l=1}^{L}\}$ are the word embeddings which shares the same context with all the attributes. Unless specified otherwise, we use $m=1$. The text encoder $f_T$ generates $\Delta v^i = f_T(\bar{Y}_i, \theta_{f_{T}})$ where $\bar{Y}_i = \{t_{SOS}, P_A^i, t_{EOS}\}$, were $t_{SOS}$ and $t_{EOS}$ are start and end token embeddings and $\theta{f_T}$ is pre-trained weights. 



While $\Delta v$ represents style, it can not be directly fed to StyleGAN as it is not compatible with its latent space. To transform $\Delta v$ to $\Delta w$, we use linear mappers ~\cite{karras2019style}: $\Delta w = M(\Delta v)$. 
Given a random latent vector $z$ as an input, a pre-trained mapping network $f_{Map}$ aims to get a latent code $w$. These latent codes $w$ are split into three groups (coarse $w_c$, middle $w_m$ and fine $w_f$) in the StyleCLIP~\cite{patashnik2021styleclip}. The Style Mapper consists of three sub-networks for coarse, middle, and fine features, and each one consists of single linear layers. The style mapper takes the output of the CLIP text encoder ($\Delta v$) and latent code $w$ and outputs editing direction $\Delta w$ for StyleGAN.
The three levels of Style Mapper $M$ can be formulated as:

\begin{align*}
\mathbf{M}(w^i, \Delta v^i) = (\mathbf{M}_c(W_c^i, \Delta v^i), \mathbf{M}_m(W_m^i, \Delta v^i),  \mathbf{M}_f(W_f^i, \Delta v^i))
\end{align*}

$\hat{w}^i = w^i + \mathbf{M}(w^i, \Delta v^i)$. The  $\hat{w}$ are passed as a input to the $G(.)$, which generates multiplane image (MPI) representation $D = \{C, \{\alpha_1 \cdots \alpha_L\}\}$, these MPIs $D$ along with the target camera pose $p_t$ are feed in the MPI Renderer $R$ to get the final generated image through generator $G_t$:
${I_{p_t}^i} = R(G_t(\hat{{w}}^i, \theta), p_t)$

Our method is data-free and extremely efficient to train. Since it is language-driven and efficient, it can be used on the fly to add and edit arbitrary new attributes. Moreover, a single set of mappers is sufficient for editing a large number of attributes by only adding additional style tokens.

\begin{figure*}[t]
    \centering
    \includegraphics[width=1\textwidth]{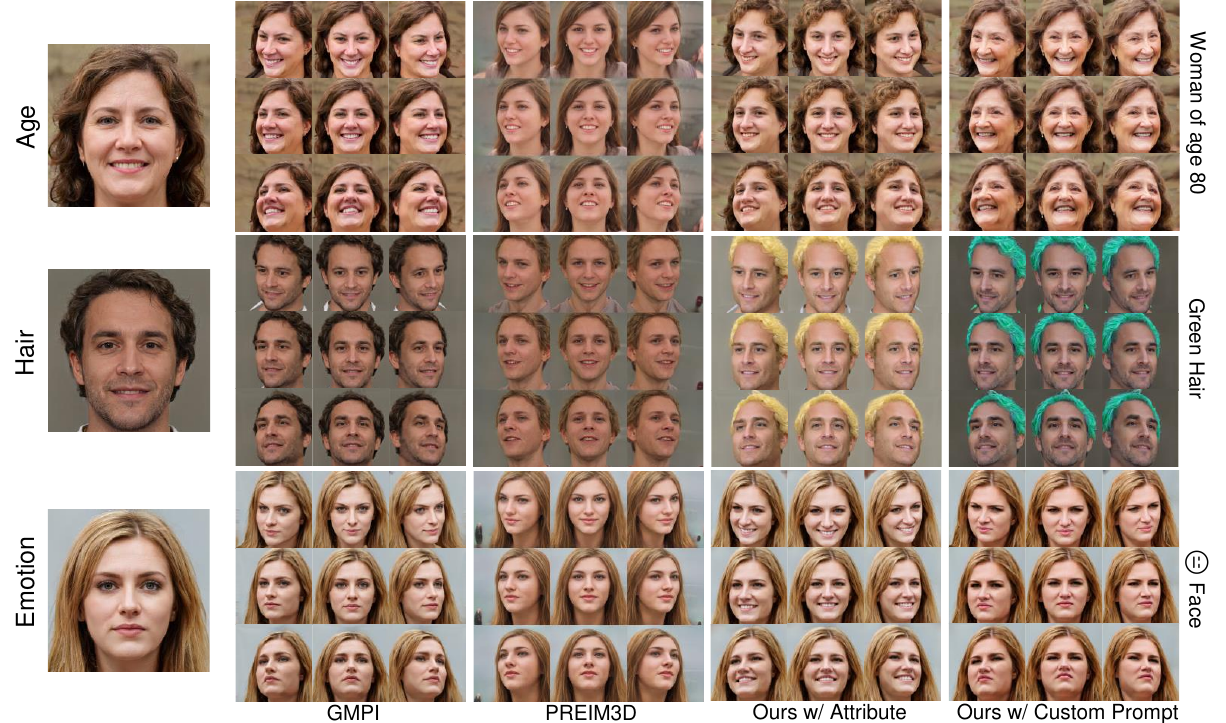}
    \caption{Our method's performance in face editing is compared qualitatively with both (GMPI~\cite{zhao2022generative}) and state-of-the-art PREIM3D~\cite{li2023preim3d} across various camera angles and attributes. The following attributes were used for comparison: young for age, blond for hair color, and happy for emotion. Additionally, to showcase our method's ability to enable the editing of novel attributes, the final column presents results obtained from custom prompts. Our method not only accurately maintains camera poses about GMPI but also demonstrates superior identity preservation and editing capability in comparison to PREIM3D.}
    \label{fig:main}
    \vspace{-20pt}
\end{figure*}
\noindent \textbf{Directional CLIP Loss for Attribute Editing. }
A simple method for guiding a generated image $I_{p_t}^i$ with the textual prompt $P_A^i$ is to align it with a target text prompt's semantic to use a CLIP-based image manipulation approach \cite{patashnik2021styleclip}. This approach involves minimizing a global clip loss function that is formulated as a way to achieve this alignment. However, such a global clip loss led to low diversity and corrupt outputs. 
To address these issues, we utilize enhanced Directional CLIP loss $L_{dclip}$~\cite{yu2022towards}, which differs from the approach used in Gal et al. \cite{Gal2021StyleGANNADACD} as it uses attribute-specific prompts and generates multi-view images on-the-fly, rather than relying on fixed, manually designed prompts and single-view images. For a given latent code $w^i$, we compute the direction of the source and target image pair at two different camera poses $p_{t_1}$ and $p_{t_2}$, 

\begin{align*}
{\Delta I_i} = \dfrac{f_I(R(f_{G_t}(\hat{w}^i), p_{t_{1}}))}{\|f_I(R(f_{G_t}(\hat{w}^i), p_{t_{1}}))\|_2} - \dfrac{f_I(R(f_{G_o}({w}^i), p_{t_{2}}))}{\|f_I(R(f_{G_o}({w}^i), p_{t_{2}}))\|_2},
\end{align*}

where $f_{G_o}$ is the original GMPI generator and $f_I$ is the image encoder of CLIP. To find attribute-specific adaptation direction, 
\begin{align}
\Delta T_i = \dfrac{f_T(P_A^i)}{\|f_T(P_A^i)\|} - 
\dfrac{f_T(t_{src})}{\|f_T(t_{src})\|},
\end{align}

where $t_{src}$ represents the semantic text of the image $I_{p_t}$. Since we are working with faces, we simply set $t_{src}$ as ``face".  The enhanced Directional CLIP loss $L_{dclip}$ is:
\begin{align}
\label{eq:Dclip}
\mathcal{L}_{dclip} = \mathbb{E}_{ w^i \in W} \sum_{i=1}^K(1 - \frac{\Delta I_ \cdot \Delta T_i}{|\Delta I_i| |\Delta T_i|} ),
\end{align}
where K is the number attribute in each batch. This loss constrains the direction of the different view images pair $\Delta I_i$  with an attribute-specific image direction $\Delta T_i$. 

\noindent\textbf{Contrastive Learning for Simultaneous Learning of Multiple Style Tokens} Our approach employs $n$ distinct style tokens, each dedicated to representing a unique attribute. Since these tokens share a style mapper, the style tokens can converge to a common orthogonal point in the text embedding space, which is common to all facial attributes. To prevent this, we introduce a novel \textit{style token contrastive loss} or $\mathcal{L}_{sc}$. This loss works by minimizing the similarity learned style tokens. Specifically, for a set of $n$ attribute-specific style tokens $\{P^1_A, P^2_A \cdots P^n_A\}$, the \textit{style token contrastive loss} is defined as,
\setlength{\abovedisplayskip}{3pt}
\begin{align}
\label{eq:const}
\mathcal{L}_{\text{sc}} = \sum_{i=1}^K \sum_{j=1, {i \neq j}}^{K-1} \text{sim}\left(  \frac{f_T(P^i_A)}{\|f_T(P^i_A)\|}, \frac{f_T(P^j_A)}{\|f_T(P^j_A)\|}  \right),
\end{align}
\setlength{\abovedisplayskip}{3pt}

where $\text{sim}(\cdot)$ represents cosine similarity and $E_T(\cdot)$ is CLIP text encoder. As our ablations demonstrate, this loss plays a pivotal role in enabling the simultaneous learning of multiple style tokens specific to each attribute. The total LAE loss is expressed as:
$$\mathcal{L_{\text{LAE}}} = \lambda_{D\text{clip}}\mathcal{L}_{D\text{dclip}} + \lambda_{\text{sc}}\mathcal{L}_{\text{sc}},$$
where $\lambda_{\text{dclip}}$ and $\lambda_{\text{sc}}$ denote the respective weights for each loss term.

\subsection{3D-aware Identity and Pose Preservation}
\label{sec:preservation_loss}

Given the latent code $\hat{w}$, our goal is to generate 3D-aware images with the different attribute prompt $P_{A}$ from a single model. To address this objective, we examine the following questions. First, How to preserve the identity of the image across different camera poses $p_{t}$ and different attributes. Second, How to prevent excessive deviation of $\alpha-$maps during training, particularly for attributes like expression and hairstyle?

The first issue is addressed by enforcing consistency in the multi-view output across different attributes and camera poses.
Finally, $\alpha-$maps preserving regularization is introduced to avoid the drastic change in the alpha maps during attribute editing. To preserve identity and pose, we utilize a few 3D-aware losses.
The textual loss ensures the final generated image $I_{p_t}$ aligns with the text prompts and is realistic, i.e., artifact-free. However, it does not constrain the identity of the image concerning the original image generated from $G_o$ and the generated image with the different camera poses and different attributes. 

\label{sec:efficiency}
\begin{table}[t]
    \centering
    \caption{We compare our method with the 3D-Inv  \cite{lin20223d},  PixelNeRF   \cite{cai2022pix2nerf} and PRIEM3D \cite{li2023preim3d}  for attribute altering (AA) and attribute dependency (AD) metrics across various attributes following PREIM3D’s protocol. Our method significantly outperforms in both measures. Here "NA" denotes that the model doesn't have a trained classifier to edit the particular attribute, whereas our model can generate any novel attribute.\vspace{-5pt}}. 
    \begin{tabular}{llccccccc}
    \toprule
        & Method    & Smile & Age & Makeup & Male & GrayHair & Lipstick    & BowlCut \\ 
        \midrule

         &3D-Inv \cite{lin20223d}     & 1.49  & 1.41  & 1.58   & 1.49  & 1.60  & 1.51  & \textbf{-} \\
         
         &Pixel2NeRF  \cite{cai2022pix2nerf}    & 1.47  & 1.42  & 1.64  & 1.55  & 1.63  & 1.56  & \textbf{-} \\
         
        AA {\color{green}($\uparrow$)} & PRIEM3D \cite{li2023preim3d}  &  1.51    & 1.54  & 1.74    & 1.62  & 1.71 & 1.62 & NA  \\

         &\cellcolor{cyan!25} \textbf{Ours}      &\cellcolor{cyan!25}  \textbf{1.69}  & \cellcolor{cyan!25} \textbf{1.62}  & \cellcolor{cyan!25} \textbf{1.86}   & \cellcolor{cyan!25} \textbf{1.70}  & \cellcolor{cyan!25} \textbf{1.76}  & \cellcolor{cyan!25} \textbf{1.71}  & \cellcolor{cyan!25} \textbf{1.66} \\
         
         \bottomrule

         &3D-Inv  \cite{lin20223d}     & 0.56  & 0.94  & 1.11   & 1.03  & 0.78  & 0.80  & \textbf{-} \\
         
         &Pixel2NeRF   \cite{cai2022pix2nerf}    & 0.57  & 1.23  & 0.92   & 0.99  & 0.61  & 0.74  & \textbf{-} \\
         
        AD {\color{red}($\downarrow$)}&PRIEM3D \cite{li2023preim3d}  &  0.49    & 0.82  & \textbf{0.88} & 0.91  & 0.63 & 0.75    & NA   \\
         &\cellcolor{cyan!25} Ours      & \cellcolor{cyan!25}  \textbf{0.42} &\cellcolor{cyan!25}  \textbf{0.71}  & \cellcolor{cyan!25} 0.89  & \cellcolor{cyan!25} \textbf{0.77}  & \cellcolor{cyan!25} \textbf{0.58}  & \cellcolor{cyan!25} \textbf{0.73}  & \cellcolor{cyan!25} \textbf{0.68}    \\ 
        \bottomrule
    \end{tabular}

    \label{tab:comparison}
    \vspace{-20pt}
\end{table}
\noindent \textbf{Identity Preservation. }
To ensure identity consistency across different camera poses and before and after the attribute change, identity loss is utilized. The identity loss aims to preserve the identity of original and modified images at a fixed front camera pose $p_o$. It is defined as follows:
\begin{align*}
\mathcal{L}_{id} = 1 - \cos(AF(R(f_{G_t}(\hat{w}), p_o)),  R(f_{G_o}(w), p_o)),
\end{align*}
where the first term is attribute modified image at the camera, pose $p_t$ and the second term is the unmodified image at the same camera pose, $\cos(\cdot)$ is the cosine similarity, and $AF(\cdot)$ is a pre-trained ArcFAce Network \cite{Deng2018ArcFaceAA} for face recognition. Here, we set the $p_t$ as the frontal angle. 

As we are using the $\mathcal{L}_{sc}$, we need to avoid the learnable prompts $P_A^i$ to deviating too much and to maintain the 3D-consistency across with camera poses $p_t$ and different attributes. To this end, we introduce identity loss for view consistency $\mathcal{L}_{idvc}$ loss. This loss aims to minimize identity differences at different poses. Hence, it samples attribute modified and unmodified images at different poses and minimizes identity loss between them. The camera poses are sampled randomly. The IDVC loss is defined as follows,
\begin{align*}
\mathcal{L}_{idvc} = \sum_{i=1}^K \sum_{j=1, {i \neq j}}^{K-1} 1 -  \cos(AF(R(G_t(\hat{w}^i), p_{t_1}), R(G_t(\hat{w}^j), p_{t_2}))),
\end{align*}
where $p_{t_1}$ and $p_{t_2}$ are two different camera poses sampled randomly. $\mathcal{L}_{idvc}$ ensures the identity consistency between different camera poses and different text prompts.  

\noindent \textbf{Camera Pose Preservation. }
To preserve the camera pose of the attribute-modified image, we constrain the latent code $\hat{w}^i$ to remain close to the initial value $w^i$, 
\begin{align*}
\mathcal{L}_{latent} = \|W - M(w, \Delta v)\|_2.
\end{align*}

Furthermore, as our $\alpha$-branch is also learnable, to avoid the drastic change in the alpha maps and thereby camera poses, we use $L_2$ norm at the manipulation steps  $\mathcal{L}_{\alpha} = \| H(\hat{W})\|_2$, where $H$ is the learnable alpha branch of Generator $f_{G_t}$. Overall, the preservation losses are as follows:
$$
\mathcal{L}_{P} = \lambda_{id}\mathcal{L}_{id} + \lambda_{idvc} \mathcal{L}_{idvc} + \lambda_{latent} \mathcal{L}_{latent} + \lambda_{\alpha} \mathcal{L}_{\alpha},
$$
where $\lambda_{id}$, $\lambda_{idvc}$, $\lambda_{latent}$  and $\lambda_{alpha}$ are hyperparameters for each term. The overall loss is, 
$\mathcal{L}_{total} = \mathcal{L}_{T} + \mathcal{L}_{P}.$

\begin{table*}[t]
  \caption{Comparison of depth and pose accuracy between our method and the state-of-the-art 3D generation method. Our method demonstrates depth and pose accuracy close to that of the baseline state-of-the-art 3D generation method. \vspace{-0.2cm}}
  \label{tab:quantative_1}
  \centering
  \resizebox{0.7\linewidth}{!}{ %
  \begin{tabular}{lcccccccc}
    \toprule
    &&   \multicolumn{3}{c}{\textbf{Depth} ($\downarrow$)} &&  \multicolumn{3}{c}{\textbf{Poses} ($\downarrow$)}      \\
    \cmidrule(r){3-5} \cmidrule(r){7-9}
    \textbf{Model}    && Smile   & Makeup & Age  && Smile   & Makeup & Age \\
    \midrule
    GMPI \cite{zhao2022generative} &&  0.49   & 0.49 & 0.49  && 0.00040 & 0.00040 & 0.00040    \\
    \cellcolor{cyan!25} Ours &\cellcolor{cyan!25}&\cellcolor{cyan!25}  0.51 & \cellcolor{cyan!25}0.50  & \cellcolor{cyan!25}0.52  &\cellcolor{cyan!25}&\cellcolor{cyan!25} 0.00042 &\cellcolor{cyan!25} 0.00046 &\cellcolor{cyan!25} 0.00043    \\ 
    \midrule
    EG3D \cite{chan2022efficient} &&  0.31 & 0.31 & 0.31   &&   0.00050 & 0.00050 & 0.00050  \\
    \cellcolor{cyan!25} Ours  &\cellcolor{cyan!25}& \cellcolor{cyan!25} 0.33    & \cellcolor{cyan!25}0.31  & \cellcolor{cyan!25}0.33 &\cellcolor{cyan!25}& \cellcolor{cyan!25} 0.00053 & \cellcolor{cyan!25}0.00054 &\cellcolor{cyan!25} 0.00051  \\
    \bottomrule
  \end{tabular}
  }
  \vspace{-15pt}
\end{table*}

\subsection{Efficiency}
In addition to enabling the editing of novel attributes defined through language, our method is significantly more efficient compared with other methods. Firstly, our approach swiftly learns to edit new attributes with a small training time (4 to 8 minutes), a notable contrast to PREIM3D's requirement of several hours of training alongside the necessity of a pre-trained attribute-specific classifier. Secondly, despite employing a language encoder, our method's inference time is comparable to that of PREIM3D. This efficiency stems from the fixed nature of the prompt post-training, allowing us to compute prompt features once and reuse them efficiently.
Finally, our method introduces a sharing mechanism for style mappers across multiple attributes, delivering benefits in both storage and computation.

\begin{figure}
    \centering

    \begin{minipage}[b]{0.4\columnwidth}
        \centering
         \includegraphics[width=0.83\columnwidth]{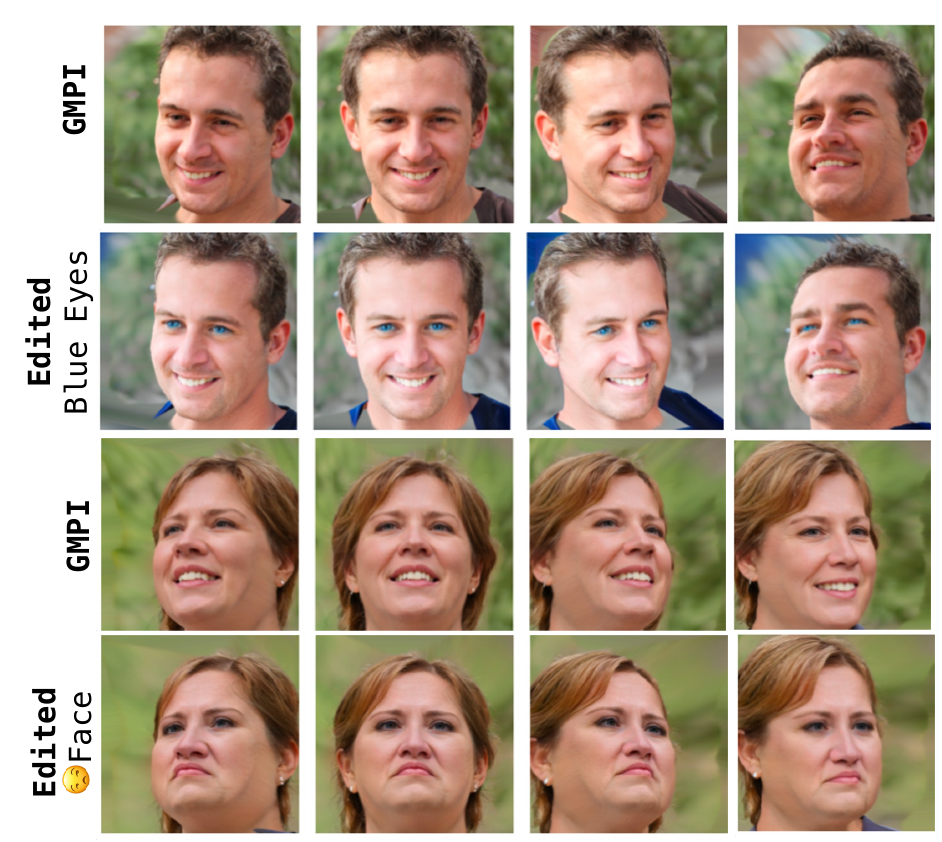}
        \subcaption{}
        \vspace{-10pt}
        \label{fig:plug}
    \end{minipage}%
    \begin{minipage}[b]{0.6\columnwidth}
        \centering
         \includegraphics[width=0.87\columnwidth]{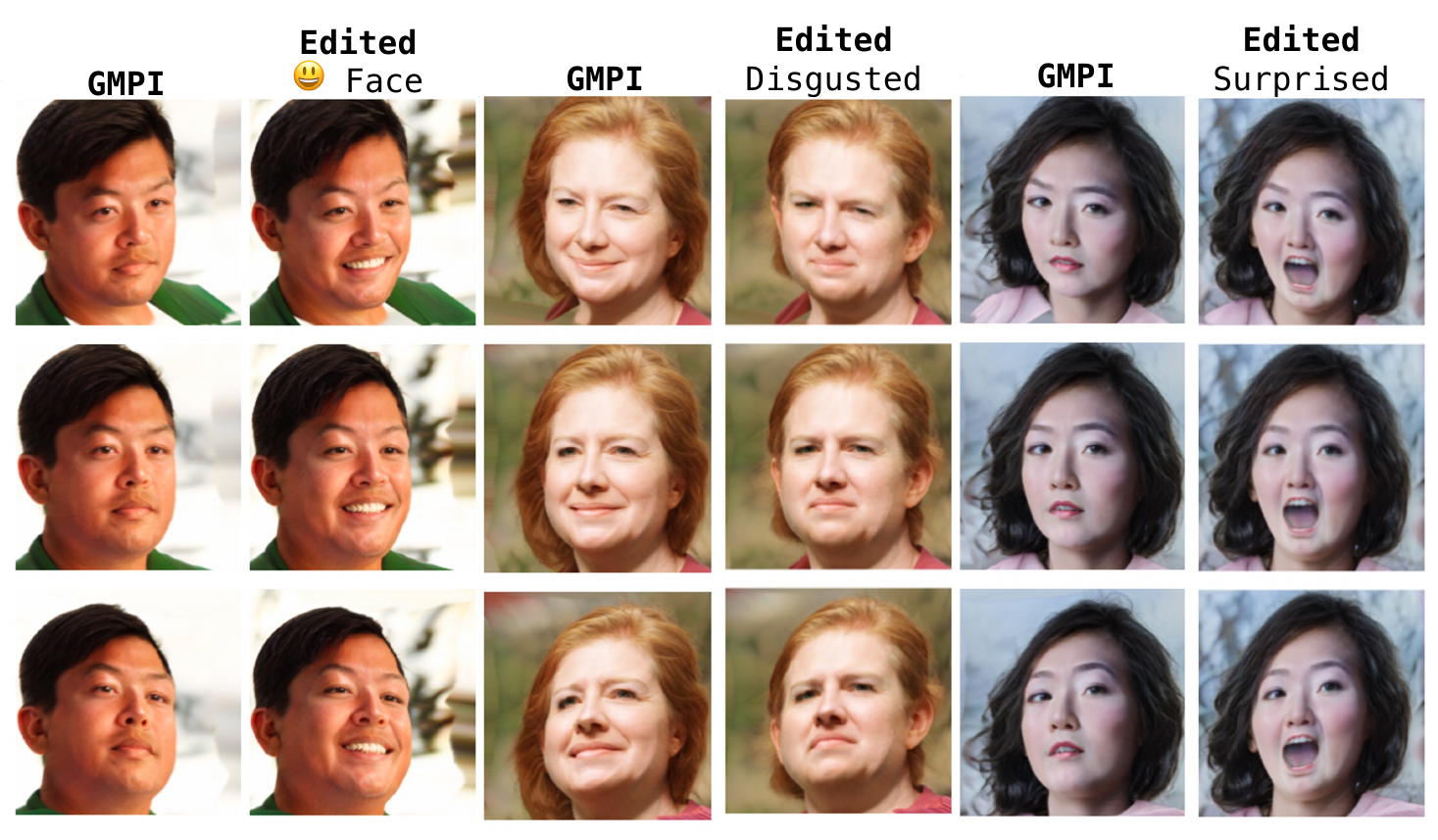}
        \subcaption{}
        \vspace{-10pt}
        \label{fig:plug}
    \end{minipage}%
    \caption{(a) Comparison between our method and GMPI  in terms of their respective abilities to maintain camera pose and identity preservation. (b) Results of editing real images for three distinct attributes. }
    \label{fig:ablation_real}
    \vspace{-15pt}
\end{figure}
\section{Experiment}
\label{sec:experiment}
\noindent\textbf{Implementation Details. }
\label{sec:implementation_details}
For the bulk of our experiments, we used a pre-trained 3D generative model GMPI~\cite{zhao2022generative} as our base 3D GAN  and kept it frozen except for the alpha maps branch. To learn the attribute-specific prompt, we use a pre-trained CLIP~\cite{radford2021learning} model. CLIP's text encoder is utilized for both training and inference, and its image encoder is used during training only. To train our LAE module, we set $\lambda_{dclip}=1.0$, $\lambda_{sc}=0.8$, $\lambda_{id}=0.8$, $\lambda_{idvc}=0.5$, $\lambda_{latent}=0.5$ and $\lambda_{\alpha}=0.5$. We optimize using Adam \cite{kingma2014adam} with a learning rate of $0.001$ and parameters $\beta_{1} = 0.9$ and $\beta_{2} = 0.95$. Similar to GMPI~\cite{zhao2022generative}, we use $32$ planes for training and $96$ planes for inference and set near and far depth for MPI as $0.95/1.12$, use the depth normalization as \cite{zhao2022generative}. To further evaluate, we integrate our method to other 3D generative model like EG3D~\cite{chan2022efficient}, StyleNeRF~\cite{gu2021stylenerf} and CIP3D~\cite{zhou2021cips}. Our method is extremely efficient in terms of both computation and space requirements, as it only requires style tokens and style mappers to be trained and stored.

\subsection{Quantitative Results}

We compare our method with the existing state-of-the-art 3D editing models against the multiple attributes shown in Table~\ref{tab:comparison}. Similar to \cite{li2023preim3d}, we use Attribute Altering(AA) and Attribute Dependency(AD) metrics. Attribute altering (AA) measures the change of the desired attribute according to the given text prompt, and attribute dependency (AD) measures the changes of other attributes given while changing a particular attribute. We can see a consistent improvement across all the attributes in Table~\ref{tab:comparison}, which shows our model had better editing ability than other SOTA 3D editing models without affecting the other attributes. 

\begin{figure}
    \centering
    \includegraphics[width=1.\textwidth]
    {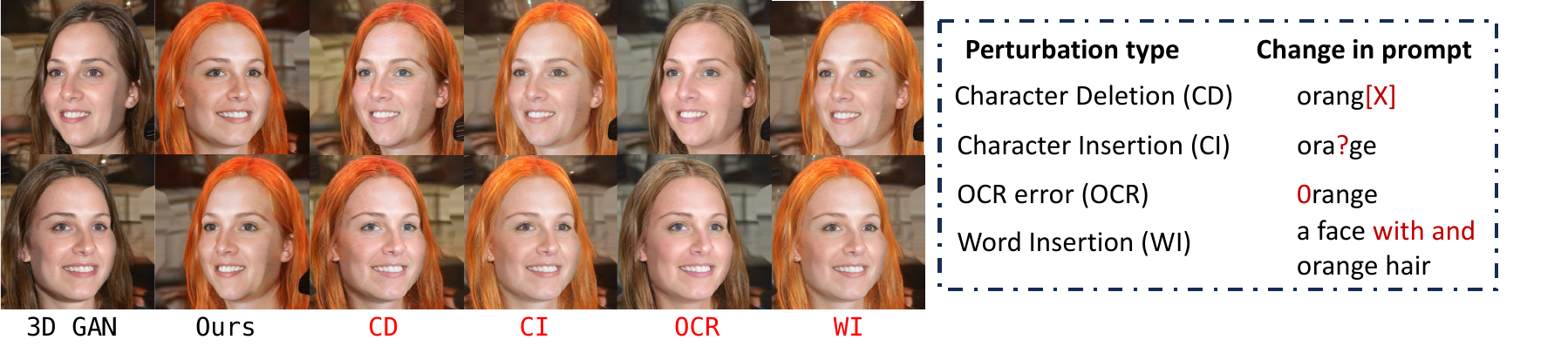}\vspace{-12pt}
    \caption{Robustness of our model to text corruptions. We introduce standard text corruptions in the text prompt to edit for the orange hair attribute. Our model is robust overall, except when the perturbation alters the keywords of the prompt. }
    \label{fig:robustness}
    \vspace{-20pt}
\end{figure}
As GMPI \cite{zhao2022generative} and EG3D \cite{chan2022efficient} lacks attribute editing capabilities, we present in Table~\ref{tab:quantative_1} a single averaged value across 1000 images. In contrast, our method calculates this value individually across three different attributes (smile, makeup, and age) using 1000 images for each attribute, reporting the average value. In a comparison of the depth and pose accuracy between GMPI \cite{zhao2022generative}, EG3D \cite{chan2022efficient} and our method. As depicted in the table, our method faithfully preserves depth and pose accuracy across attributes. For instance, concerning smile attribute editing, our method exhibits a depth accuracy of 0.51, marginally lower than GMPI's 0.49 (lower is better) while using GMPI backbone, when integrating the LAE Module with the EG3D we can observe a similar pattern. Similar observations are evident for makeup and age attributes. These findings highlight our method's capability to maintain 3D geometry while facilitating attribute editing.

\subsection{Qualitative Results}

Firstly, to demonstrate the efficacy of our method, we showcase qualitative results and compare them with the GMPI and the state-of-the-art PREIM3D~\cite{li2023preim3d} in Figure~\ref{fig:main}. Following the approach outlined in~\cite{li2023preim3d}, we uniformly sampled 9 images with yaw angles ranging from -30\textdegree{} to 30\textdegree{} and pitch angles between -20\textdegree{} and 20\textdegree{}. To facilitate comparative analysis, we utilize pre-trained weights provided by GMPI~\cite{zhao2022generative} and PREIM3D~\cite{li2023preim3d} and replicate their results.

Our method excels in preserving identity compared with the generated images by GMPI and demonstrates superior 3D consistency across varying camera angles compared to PREIM3D. For instance, when altering hair color in the second row, our method faithfully reconstructs the subject's identity. Similarly, our method also preserves camera poses. 
Moreover, our model exhibits flexibility as it can alter attributes specified by natural language. To showcase this adaptability, the last column in Figure~\ref{fig:main} displays results obtained using arbitrary natural language prompts. Specifically, to edit a face with a new attribute, PREIM3D necessitates a pre-trained attribute classifier and several hours of training. In contrast, our method achieves the same with a few minutes of training. We illustrate this efficiency for prompts such as green hair, emoji face, and aging.

\begin{figure}
    \centering

    \begin{minipage}[b]{0.6\columnwidth}
        \centering
         \includegraphics[width=1\columnwidth]{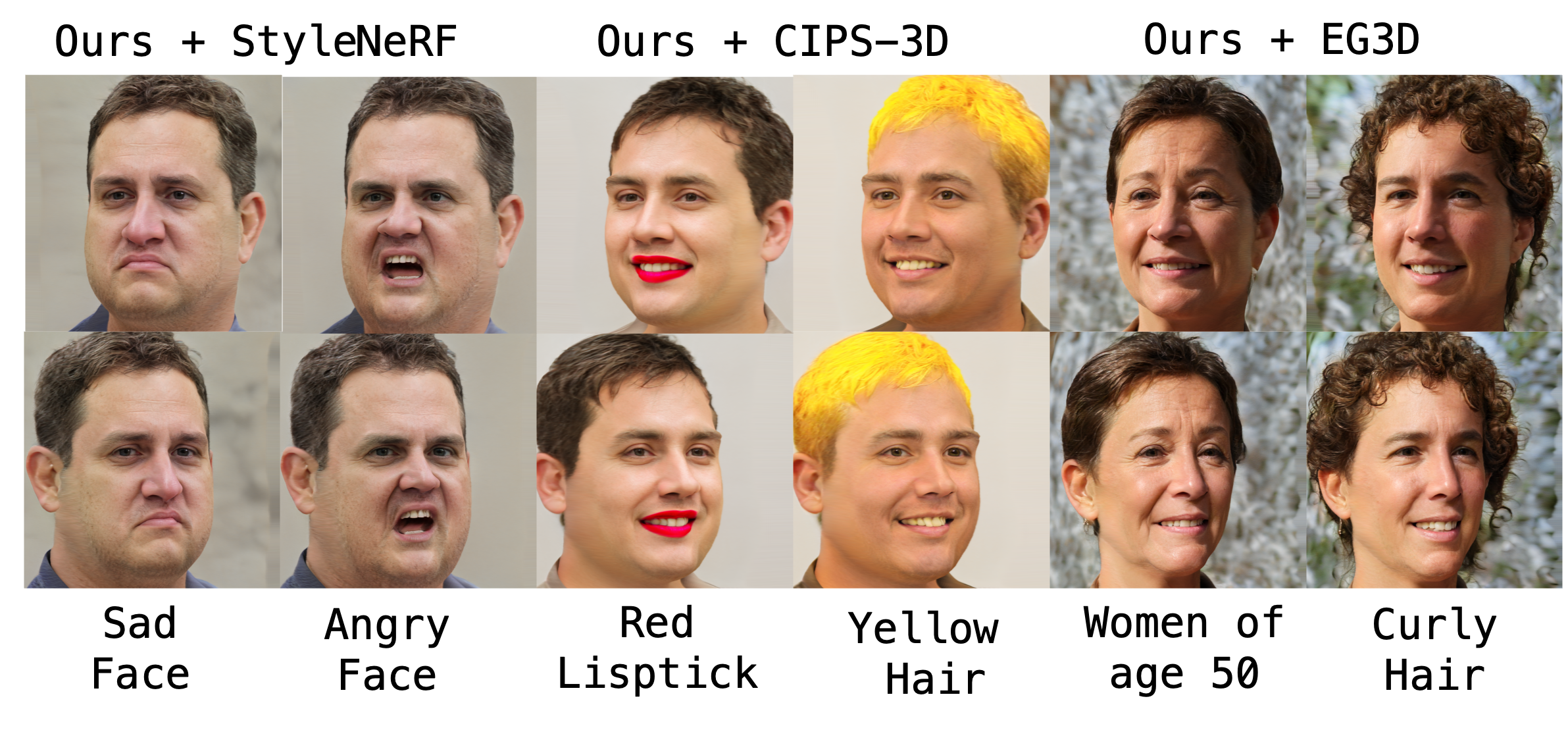}
        \subcaption{}
        \label{fig:plug}
    \end{minipage}%
    \begin{minipage}[b]{0.34\columnwidth}
        \centering
        \resizebox{0.75\columnwidth}{!}{
    \begin{tabular}{lccc}
    \toprule
        Method & FID {\color{red}($\downarrow$)}  & KID {\color{red}($\downarrow$)} \\
        \toprule
         GMPI   & 25.91 & 0.074   \\
         + Ours   & 26.57 & 0.076   \\
         \midrule
         StyleNeRF   & 22.12 & 0.039  \\ 
         + Ours   & 24.49 & 0.042  \\ 
         \midrule
         CIPS3D   & 17.89   & 0.031  \\
         + Ours   & 18.01   & 0.033  \\
           \midrule
         EG3D   & 13.69 & 0.018   \\
         + Ours   & 15.12 & 0.021   \\
        \bottomrule
    \end{tabular}}
        \subcaption{ }
        \label{tab:fid}
    \end{minipage}
     \vspace{-7pt}
    \caption{(a) Qualitative analysis by plugging the LAE Module into SOTA 3D Generation Model, such as StyleNeRF  \cite{gu2021stylenerf}, CIPS-3D  \cite{zhou2021cips}, and EG3D \cite{chan2022efficient}. (b) Quantitative Ablation for image quality performance after integrating our proposed LAE Module. }
    \label{fig:plug-and-play}
    \vspace{-20pt}
\end{figure}
To showcase the efficacy of our method in preserving both identity and camera poses, we conduct a comparison with GMPI in Figure~\ref{fig:ablation_real}. This comparison spans two attributes observed across four randomly selected camera angles. The figure demonstrates how our method handles the editing of intricate and challenging attributes, such as specific emojis or blue eye colors, without compromising camera angles or altering the subject's identity.
Further, we use CelebA-HQ~\cite{karras2017progressive} to illustrate our method's ability to edit real images. We first invert images to obtain their latent code using E4E~\cite{tov2021designing}. Subsequently, these images are edited using our method, and the results of this experiment are shown in Figure~\ref{fig:ablation_real}(b). Our method maintains the identity and preserves the camera pose while successfully altering attributes. It's essential to note that the preservation quality of our method is inherently constrained by GMPI.

\subsection{Analysis}
\subsubsection{Plug-and-play}

To further assess the plug-and-play capabilities of our LAE Module, we integrate it with various state-of-the-art 3D generation methods (StyleNeRF  \cite{gu2021stylenerf}, CIPS-3D  \cite{zhou2021cips}, and EG3D \cite{chan2022efficient}). Figure~\ref{fig:plug-and-play} illustrates the efficacy of our LAE Module in attribute editing across different 3D generation methods. The attribute-specific prompts $\textbf{P}_{A}^{i}$ are learned to find the editing direction in $W$ space of the above 3D models and to have multi-view consistency and identity preservation which uses our proposed $L_{idvc}$ and $L_{latent}$ loss. Table~\ref{fig:plug-and-play}(b) demonstrates the image quality performance of our model by integrating our LAE module with the state-of-art 3D Generation model, using metrics such as FID \cite{heusel2017gans} and KID \cite{binkowski2018demystifying}. Our method adds editing capabilities to the state-of-the-art 3D generative model without comprising the generation quality of the generated images with particular attributes.

\subsubsection{Robustness and biasness}

To understand the robustness of our model with imprecise prompts, We use four standard text perturbations \cite{qiu2023benchmarking} (character deletion (CD), word insertion (WI), OCR, and Back translation (BT)) to edit for orange hair attribute as presented in Fig~\ref{fig:robustness}. Our model is robust overall, except when the perturbation alters the keyword, such as changing ``Orange" to ``(\textcolor{violet}{0})range", exhibiting limitations of the CLIP encoding.

\begin{figure}
    \centering
    \begin{minipage}[b]{0.33\columnwidth}
        \centering
         \includegraphics[width=0.83\columnwidth]{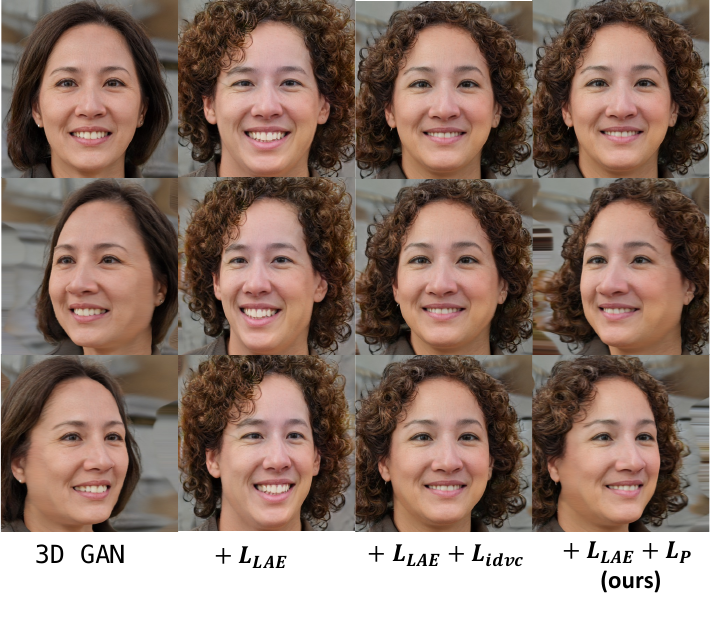}
        \label{fig:plug}
    \end{minipage}
    \begin{minipage}[b]{0.58\columnwidth}
        \centering
        \resizebox{1\columnwidth}{!}{
    \begin{tabular}{lcccc}
    \toprule
        Method & $\text{ID}_{(0-10)}$ ({\color{green} $\uparrow$}) & ID (10-30) ({\color{green} $\uparrow$}) & Depth {\color{red}($\downarrow$)} & Poses {\color{red}($\downarrow$)} \\
        \toprule
         3D-GAN   & 0.73 & 0.70 & 0.53 & 0.0004 \\
         \midrule
         $+L_{LAE}$  & 0.66 & 0.65 & 0.61 & 0.0006 \\ 
         \midrule
         $+L_{LAE}+L_{idvc}$   & 0.71   & 0.68 & 0.60 & 0.0006 \\
           \midrule
         $+L_{total}$ (Ours)   & 0.72 & 0.70 & 0.54 & 0.0004 \\
        \bottomrule
    \end{tabular}
    }
     \vspace{22pt}
    \end{minipage}
   \vspace{-15pt}
    \caption{Qualitative and Quantitative ablation of losses on ID, depth, and pose. }
    \label{fig:ablation_study}
    \vspace{-15pt}
\end{figure}

To delve deeper into understanding the influence of attributes on each other, such as how changing "blue eyes" may affect "skin tone" as shown in row 2 of Figure.~\ref{fig:ablation_real}(a), we explored the bias present in the CLIP model.   Our method finds semantic direction in the latent space of 3D-aware GANs via. text, encoded by CLIP. Attribute editing accuracy and control rely on CLIP's encoding, potentially inheriting biases, noted in previous works\cite{berg2022prompt}. To understand the biases of the clip, we performed a simple experiment by encoding ten variations of prompts for white, brown, and black faces and blue eyes with CLIP and calculating the distance of the faces with blue eyes. We found that the encoding of the white face is closest to the blue eyes: white $0.035$, brown $0.051$, and black $0.065$.

\subsection{Ablation Study}

In Figure~\ref{fig:ablation_study}(a), highlights the impact of introducing proposed losses and the LAE Module into the 3D generation method for the attribute "curly hair". Integrating the LAE Module with $L_{LAE}$ loss brings about changes in the desired attribute, emphasizing the significance of using $L_{LAE}$ loss during the training of the LAE Module. However, despite altering the desired attributes, the LAE module falls short of preserving the identity and poses of the edited images. This issue is addressed by introducing $L_{id}+L_{idvc}$, resulting in images with desired attributes while maintaining a similar identity to the original image, though poses are still not adequately preserved. Our final approach, employing $L_{total}$, effectively preserves poses, identity, and generates 3D-aware edited images

Table~\ref{fig:ablation_study}(b) presents a quantitative evaluation of the effect of losses on preserving identity and different camera poses across the "curly hair" attribute. Following prior works \cite{zhao2022generative, chan2022efficient}, the mean ArcFace Similarity score is assessed across various generated images and edited faces under random camera poses.  Integrating the LAE module with $L_{LAE}$ results in an ID score of 0.66 for camera poses ranging from 0 to 10 and 0.65 for poses from 10 to 30, while depth and poses exhibit scores around 0.61 and 0.0006, respectively. With training using $L_{total}$, a consistent enhancement is observed across all metrics, with ID score improving from 0.66 to 0.72, depth from 0.61 to 0.54, and poses from 0.0004 to 0.0004.

\subsection{Conclusion}

In this paper, we present an efficient pipeline for editing 3D-aware and view-consistent facial image attributes specified through textual prompts. Our method comprises a text-driven Latent Attribute Editor (LAE) alongside a 3D GAN. The LAE integrates learned style tokens and a style mapper, leveraging a pre-trained CLIP model to find semantic editing directions within 3D GAN latent space. The text-driven approach and utilization of style tokens render our method exceptionally efficient in handling novel editing directions. Additionally, we employ several 3D-aware identities and pose preservation losses to ensure view consistency in the generated images. Our method's effectiveness is validated through a comprehensive array of qualitative and quantitative experiments.

%
%
\bibliographystyle{splncs04}
\bibliography{main}

\begin{thebibliography}{10}
\providecommand{\url}[1]{\texttt{#1}}
\providecommand{\urlprefix}{URL }
\providecommand{\doi}[1]{https://doi.org/#1}

\bibitem{abdal2021styleflow}
Abdal, R., Zhu, P., Mitra, N.J., Wonka, P.: Styleflow: Attribute-conditioned exploration of stylegan-generated images using conditional continuous normalizing flows. ACM Transactions on Graphics (ToG)  \textbf{40}(3),  1--21 (2021)

\bibitem{alayrac2022flamingo}
Alayrac, J.B., Donahue, J., Luc, P., Miech, A., Barr, I., Hasson, Y., Lenc, K., Mensch, A., Millican, K., Reynolds, M., et~al.: Flamingo: a visual language model for few-shot learning. NeurIPS  \textbf{35} (2022)

\bibitem{berg2022prompt}
Berg, H., Hall, S.M., Bhalgat, Y., Yang, W., Kirk, H.R., Shtedritski, A., Bain, M.: A prompt array keeps the bias away: Debiasing vision-language models with adversarial learning. arXiv preprint arXiv:2203.11933  (2022)

\bibitem{binkowski2018demystifying}
Bi{\'n}kowski, M., Sutherland, D.J., Arbel, M., Gretton, A.: Demystifying mmd gans. arXiv preprint arXiv:1801.01401  (2018)

\bibitem{brock2018large}
Brock, A., Donahue, J., Simonyan, K.: Large scale gan training for high fidelity natural image synthesis. arXiv preprint arXiv:1809.11096  (2018)

\bibitem{cai2022pix2nerf}
Cai, S., Obukhov, A., Dai, D., Van~Gool, L.: Pix2nerf: Unsupervised conditional p-gan for single image to neural radiance fields translation. In: Proceedings of the IEEE/CVF conference on computer vision and pattern recognition. pp. 3981--3990 (2022)

\bibitem{chan2022efficient}
Chan, E.R., Lin, C.Z., Chan, M.A., Nagano, K., Pan, B., De~Mello, S., Gallo, O., Guibas, L.J., Tremblay, J., Khamis, S., et~al.: Efficient geometry-aware 3d generative adversarial networks. In: Proceedings of the IEEE/CVF Conference on Computer Vision and Pattern Recognition. pp. 16123--16133 (2022)

\bibitem{chan2021pi}
Chan, E.R., Monteiro, M., Kellnhofer, P., Wu, J., Wetzstein, G.: pi-gan: Periodic implicit generative adversarial networks for 3d-aware image synthesis. In: Proceedings of the IEEE/CVF conference on computer vision and pattern recognition. pp. 5799--5809 (2021)

\bibitem{collins2020editing}
Collins, E., Bala, R., Price, B., Susstrunk, S.: Editing in style: Uncovering the local semantics of gans. In: Proceedings of the IEEE/CVF Conference on Computer Vision and Pattern Recognition. pp. 5771--5780 (2020)

\bibitem{Deng2018ArcFaceAA}
Deng, J., Guo, J., Zafeiriou, S.: Arcface: Additive angular margin loss for deep face recognition. 2019 IEEE/CVF Conference on Computer Vision and Pattern Recognition (CVPR) pp. 4685--4694 (2018)

\bibitem{gadelha20173d}
Gadelha, M., Maji, S., Wang, R.: 3d shape induction from 2d views of multiple objects. In: 2017 International Conference on 3D Vision (3DV). pp. 402--411. IEEE (2017)

\bibitem{Gal2021StyleGANNADACD}
Gal, R., Patashnik, O., Maron, H., Chechik, G., Cohen-Or, D.: Stylegan-nada: Clip-guided domain adaptation of image generators. ArXiv  \textbf{abs/2108.00946} (2021)

\bibitem{goodfellow2020generative}
Goodfellow, I., Pouget-Abadie, J., Mirza, M., Xu, B., Warde-Farley, D., Ozair, S., Courville, A., Bengio, Y.: Generative adversarial networks. Communications of the ACM  \textbf{63}(11),  139--144 (2020)

\bibitem{gu2021stylenerf}
Gu, J., Liu, L., Wang, P., Theobalt, C.: Stylenerf: A style-based 3d-aware generator for high-resolution image synthesis. arXiv preprint arXiv:2110.08985  (2021)

\bibitem{harkonen2020ganspace}
H{\"a}rk{\"o}nen, E., Hertzmann, A., Lehtinen, J., Paris, S.: Ganspace: Discovering interpretable gan controls. Advances in neural information processing systems  \textbf{33},  9841--9850 (2020)

\bibitem{henzler2019escaping}
Henzler, P., Mitra, N.J., Ritschel, T.: Escaping plato's cave: 3d shape from adversarial rendering. In: Proceedings of the IEEE/CVF International Conference on Computer Vision. pp. 9984--9993 (2019)

\bibitem{heusel2017gans}
Heusel, M., Ramsauer, H., Unterthiner, T., Nessler, B., Hochreiter, S.: Gans trained by a two time-scale update rule converge to a local nash equilibrium. Advances in neural information processing systems  \textbf{30} (2017)

\bibitem{jia2021scaling}
Jia, C., Yang, Y., Xia, Y., Chen, Y.T., Parekh, Z., Pham, H., Le, Q., Sung, Y.H., Li, Z., Duerig, T.: Scaling up visual and vision-language representation learning with noisy text supervision. In: ICML. PMLR (2021)

\bibitem{karras2017progressive}
Karras, T., Aila, T., Laine, S., Lehtinen, J.: Progressive growing of gans for improved quality, stability, and variation. arXiv preprint arXiv:1710.10196  (2017)

\bibitem{karras2020training}
Karras, T., Aittala, M., Hellsten, J., Laine, S., Lehtinen, J., Aila, T.: Training generative adversarial networks with limited data. Advances in neural information processing systems  \textbf{33},  12104--12114 (2020)

\bibitem{karras2019style}
Karras, T., Laine, S., Aila, T.: A style-based generator architecture for generative adversarial networks. In: Proceedings of the IEEE/CVF conference on computer vision and pattern recognition. pp. 4401--4410 (2019)

\bibitem{karras2020analyzing}
Karras, T., Laine, S., Aittala, M., Hellsten, J., Lehtinen, J., Aila, T.: Analyzing and improving the image quality of stylegan. In: Proceedings of the IEEE/CVF conference on computer vision and pattern recognition. pp. 8110--8119 (2020)

\bibitem{kingma2014adam}
Kingma, D.P., Ba, J.: Adam: A method for stochastic optimization. arXiv preprint arXiv:1412.6980  (2014)

\bibitem{kumar2023generative}
Kumar, A., Bhunia, A.K., Narayan, S., Cholakkal, H., Anwer, R.M., Khan, S., Yang, M.H., Khan, F.S.: Generative multiplane neural radiance for 3d-aware image generation. In: Proceedings of the IEEE/CVF International Conference on Computer Vision. pp. 7388--7398 (2023)

\bibitem{li2023preim3d}
Li, J., Li, J., Zhang, H., Liu, S., Wang, Z., Xiao, Z., Zheng, K., Zhu, J.: Preim3d: 3d consistent precise image attribute editing from a single image. In: Proceedings of the IEEE/CVF Conference on Computer Vision and Pattern Recognition. pp. 8549--8558 (2023)

\bibitem{li2022blip}
Li, J., Li, D., Xiong, C., Hoi, S.: Blip: Bootstrapping language-image pre-training for unified vision-language understanding and generation. In: ICML. PMLR (2022)

\bibitem{liao2020towards}
Liao, Y., Schwarz, K., Mescheder, L., Geiger, A.: Towards unsupervised learning of generative models for 3d controllable image synthesis. In: Proceedings of the IEEE/CVF conference on computer vision and pattern recognition. pp. 5871--5880 (2020)

\bibitem{lin20223d}
Lin, C.Z., Lindell, D.B., Chan, E.R., Wetzstein, G.: 3d gan inversion for controllable portrait image animation. arXiv preprint arXiv:2203.13441  (2022)

\bibitem{miyato2018spectral}
Miyato, T., Kataoka, T., Koyama, M., Yoshida, Y.: Spectral normalization for generative adversarial networks. arXiv preprint arXiv:1802.05957  (2018)

\bibitem{nguyen2020blockgan}
Nguyen-Phuoc, T.H., Richardt, C., Mai, L., Yang, Y., Mitra, N.: Blockgan: Learning 3d object-aware scene representations from unlabelled images. Advances in neural information processing systems  \textbf{33},  6767--6778 (2020)

\bibitem{patashnik2021styleclip}
Patashnik, O., Wu, Z., Shechtman, E., Cohen-Or, D., Lischinski, D.: Styleclip: Text-driven manipulation of stylegan imagery. In: Proceedings of the IEEE/CVF International Conference on Computer Vision. pp. 2085--2094 (2021)

\bibitem{qiu2023benchmarking}
Qiu, J., Zhu, Y., Shi, X., Wenzel, F., Tang, Z., Zhao, D., Li, B., Li, M.: Benchmarking robustness of multimodal image-text models under distribution shift. Journal of Data-centric Machine Learning Research  (2023)

\bibitem{radford2021learning}
Radford, A., Kim, J.W., Hallacy, C., Ramesh, A., Goh, G., Agarwal, S., Sastry, G., Askell, A., Mishkin, P., Clark, J., et~al.: Learning transferable visual models from natural language supervision. In: ICML. PMLR (2021)

\bibitem{radford2015unsupervised}
Radford, A., Metz, L., Chintala, S.: Unsupervised representation learning with deep convolutional generative adversarial networks. arXiv preprint arXiv:1511.06434  (2015)

\bibitem{roich2022pivotal}
Roich, D., Mokady, R., Bermano, A.H., Cohen-Or, D.: Pivotal tuning for latent-based editing of real images. ACM Transactions on graphics (TOG)  \textbf{42}(1),  1--13 (2022)

\bibitem{schwarz2020graf}
Schwarz, K., Liao, Y., Niemeyer, M., Geiger, A.: Graf: Generative radiance fields for 3d-aware image synthesis. Advances in Neural Information Processing Systems  \textbf{33},  20154--20166 (2020)

\bibitem{shen2020interpreting}
Shen, Y., Gu, J., Tang, X., Zhou, B.: Interpreting the latent space of gans for semantic face editing. In: Proceedings of the IEEE/CVF conference on computer vision and pattern recognition. pp. 9243--9252 (2020)

\bibitem{shen2020interfacegan}
Shen, Y., Yang, C., Tang, X., Zhou, B.: Interfacegan: Interpreting the disentangled face representation learned by gans. IEEE transactions on pattern analysis and machine intelligence  \textbf{44}(4),  2004--2018 (2020)

\bibitem{singh2022flava}
Singh, A., Hu, R., Goswami, V., Couairon, G., Galuba, W., Rohrbach, M., Kiela, D.: Flava: A foundational language and vision alignment model. In: Proceedings of the IEEE CVPR (2022)

\bibitem{sun2022ide}
Sun, J., Wang, X., Shi, Y., Wang, L., Wang, J., Liu, Y.: Ide-3d: Interactive disentangled editing for high-resolution 3d-aware portrait synthesis. ACM Transactions on Graphics (ToG)  \textbf{41}(6),  1--10 (2022)

\bibitem{szabo2019unsupervised}
Szab{\'o}, A., Meishvili, G., Favaro, P.: Unsupervised generative 3d shape learning from natural images. arXiv preprint arXiv:1910.00287  (2019)

\bibitem{tewari2020stylerig}
Tewari, A., Elgharib, M., Bharaj, G., Bernard, F., Seidel, H.P., P{\'e}rez, P., Zollhofer, M., Theobalt, C.: Stylerig: Rigging stylegan for 3d control over portrait images. In: Proceedings of the IEEE/CVF Conference on Computer Vision and Pattern Recognition. pp. 6142--6151 (2020)

\bibitem{tov2021designing}
Tov, O., Alaluf, Y., Nitzan, Y., Patashnik, O., Cohen-Or, D.: Designing an encoder for stylegan image manipulation. ACM Transactions on Graphics (TOG)  \textbf{40}(4),  1--14 (2021)

\bibitem{wu2016learning}
Wu, J., Zhang, C., Xue, T., Freeman, B., Tenenbaum, J.: Learning a probabilistic latent space of object shapes via 3d generative-adversarial modeling. Advances in neural information processing systems  \textbf{29} (2016)

\bibitem{wu2021stylespace}
Wu, Z., Lischinski, D., Shechtman, E.: Stylespace analysis: Disentangled controls for stylegan image generation. In: Proceedings of the IEEE/CVF Conference on Computer Vision and Pattern Recognition. pp. 12863--12872 (2021)

\bibitem{xie2023high}
Xie, J., Ouyang, H., Piao, J., Lei, C., Chen, Q.: High-fidelity 3d gan inversion by pseudo-multi-view optimization. In: Proceedings of the IEEE/CVF Conference on Computer Vision and Pattern Recognition. pp. 321--331 (2023)

\bibitem{xu20223d}
Xu, Y., Peng, S., Yang, C., Shen, Y., Zhou, B.: 3d-aware image synthesis via learning structural and textural representations. In: Proceedings of the IEEE/CVF Conference on Computer Vision and Pattern Recognition. pp. 18430--18439 (2022)

\bibitem{yao2021filip}
Yao, L., Huang, R., Hou, L., Lu, G., Niu, M., Xu, H., Liang, X., Li, Z., Jiang, X., Xu, C.: Filip: Fine-grained interactive language-image pre-training. In: ICLR (2021)

\bibitem{yu2022towards}
Yu, Y., Zhan, F., Wu, R., Zhang, J., Lu, S., Cui, M., Xie, X., Hua, X.S., Miao, C.: Towards counterfactual image manipulation via clip. In: Proceedings of the 30th ACM International Conference on Multimedia. pp. 3637--3645 (2022)

\bibitem{yuan2021florence}
Yuan, L., Chen, D., Chen, Y.L., Codella, N., Dai, X., Gao, J., Hu, H., Huang, X., Li, B., Li, C., et~al.: Florence: A new foundation model for computer vision. arXiv preprint arXiv:2111.11432  (2021)

\bibitem{zhao2022generative}
Zhao, X., Ma, F., G{\"u}era, D., Ren, Z., Schwing, A.G., Colburn, A.: Generative multiplane images: Making a 2d gan 3d-aware. In: European Conference on Computer Vision. pp. 18--35. Springer (2022)

\bibitem{zhou2021cips}
Zhou, P., Xie, L., Ni, B., Tian, Q.: Cips-3d: A 3d-aware generator of gans based on conditionally-independent pixel synthesis. arXiv preprint arXiv:2110.09788  (2021)

\bibitem{zhou2018stereo}
Zhou, T., Tucker, R., Flynn, J., Fyffe, G., Snavely, N.: Stereo magnification: Learning view synthesis using multiplane images. arXiv preprint arXiv:1805.09817  (2018)

\bibitem{zhu2018visual}
Zhu, J.Y., Zhang, Z., Zhang, C., Wu, J., Torralba, A., Tenenbaum, J., Freeman, B.: Visual object networks: Image generation with disentangled 3d representations. Advances in neural information processing systems  \textbf{31} (2018)

\end{thebibliography}
\end{document}